\documentclass{article}

\usepackage{microtype}
\usepackage{graphicx}
\usepackage{subcaption}
\usepackage{booktabs} 
\usepackage{xspace}
\usepackage{caption}

\usepackage{hyperref}

\newcommand{\ours}{$E$mbedding-perturbed $E$xploration Preference Optimization\xspace}
\newcommand{\ourshort}{$E^2$PO\xspace}
\newcommand{\supp}{Appendix\xspace}

\PassOptionsToPackage{sort&compress}{natbib}
\usepackage[preprint]{icml2026}
\setcitestyle{numbers,square,comma,sort&compress}

\usepackage{amsmath}
\usepackage{amssymb}
\usepackage{mathtools}
\usepackage{amsthm}
\usepackage{bm}      

\usepackage[capitalize,noabbrev]{cleveref}

\usepackage[utf8]{inputenc}
\usepackage{colortbl}   
\usepackage{xcolor}     
\usepackage{graphicx}   
\usepackage{siunitx}   

\usepackage{multirow}
\definecolor{graybg}{gray}{0.9}
\definecolor{teaserblue}{HTML}{1F77B4}   
\definecolor{teaserpurple}{HTML}{800080}

\theoremstyle{plain}

\theoremstyle{definition}

\theoremstyle{remark}

\usepackage[textsize=tiny]{todonotes}

\usepackage{xcolor}
\definecolor{mypurple}{RGB}{128, 0, 128}

\icmltitlerunning{E²PO: Embedding-perturbed Exploration Preference Optimization for Flow Models}

\begin{document}

\twocolumn[
\vspace{-20pt}
    \icmltitle{%
      \protect\textbf{\protect\textcolor{mypurple}{E$^2$}}PO: 
      \protect\textbf{\protect\textcolor{mypurple}{E}}mbedding-perturbed 
      \protect\textbf{\protect\textcolor{mypurple}{E}}xploration 
      Preference Optimization \\ for Flow Models
    }
 \vspace{-5pt}

    \icmlsetsymbol{equal}{*}
    \icmlsetsymbol{corr}{$\dagger$}
    \icmlsetsymbol{lead}{$\ddagger$}
    \icmlsetsymbol{researchlead}{$\mathsection$} 
    \icmlsetsymbol{done}{$\star$}
    
    \begin{icmlauthorlist}
        \icmlauthor{Sujie Hu}{equal,done,thu}
        \icmlauthor{Chubin Chen}{equal,lead,thu}
        \icmlauthor{Jiashu Zhu}{equal,researchlead,ali}
        \icmlauthor{Jiahong Wu}{corr,ali}
        \icmlauthor{Xiangxiang Chu}{ali}
        \icmlauthor{Xiu Li}{corr,thu}
    \end{icmlauthorlist}
    
    \icmlaffiliation{thu}{Tsinghua University}
    \icmlaffiliation{ali}{AMAP, Alibaba Group}
    

  \icmlkeywords{Reinforcement Learning, Diffusion Models}
  \icmlkeywords{Machine Learning, ICML}

  \vskip 0.1in

  \begin{center}
    {\includegraphics[width=\textwidth]{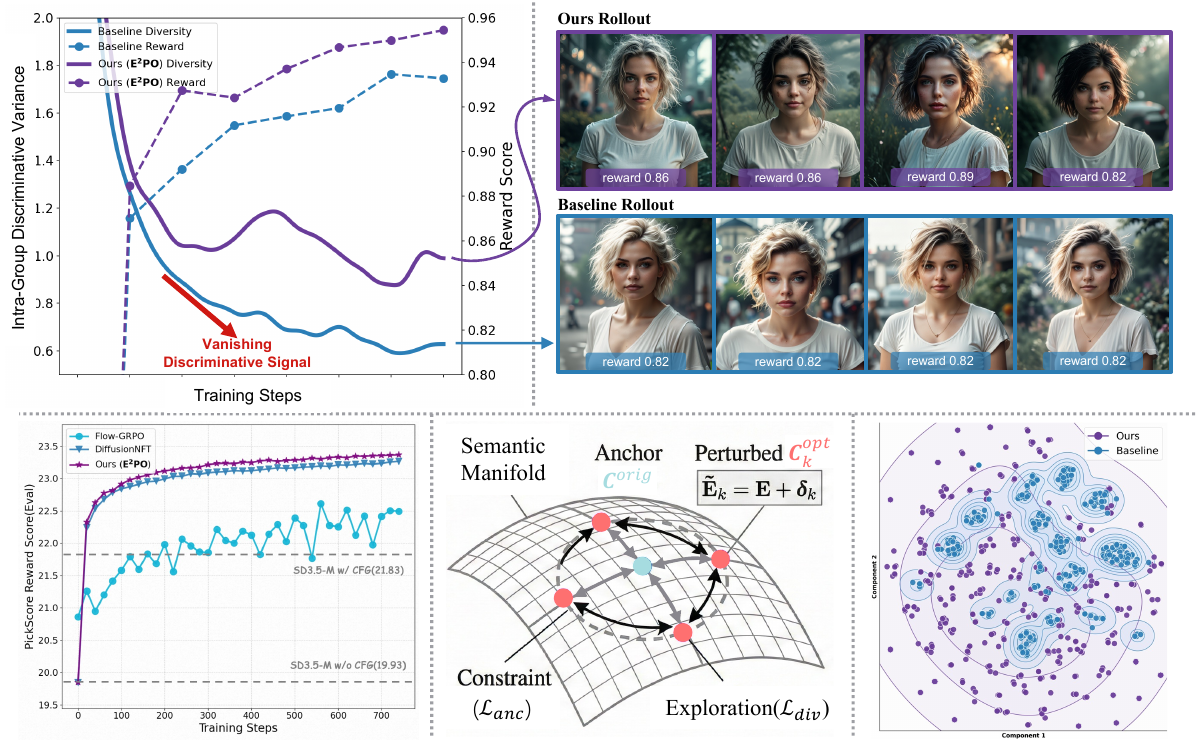}}
    
    \captionof{figure}{
\textbf{(Top)} Unlike \textcolor{teaserblue}{\textbf{baselines}} that suffer from \textit{Vanishing Discriminative Signal}, \textcolor{teaserpurple}{\textbf{\ourshort}} sustains Discriminative Variance through embedding-level perturbation. 
\textbf{(Bottom)} By injecting structured perturbations into the semantic manifold (Center), our method significantly expands the exploration space (Right) and achieves superior reward alignment (Left) compared to state-of-the-art methods.}
    \vspace{15pt}
    \label{fig:teaser}
  \end{center}

]



\printAffiliationsAndNotice{%
    $^*$Equal contribution \quad
    $^\dagger$Corresponding author \quad
    $^\ddagger$Project lead \quad
    $^\mathsection$Research lead
    $^\star$Work done during the internship at AMAP
}

\begin{abstract}

Recent advancements have established Reinforcement Learning (RL) as a pivotal paradigm for aligning generative models with human intent. 
However, group-based optimization frameworks (e.g., GRPO) face a critical limitation: \textit{the rapid decay of intra-group variance}. As the distinctiveness among samples within a group diminishes, the variance approaches zero. This eliminates the very learning signal required for optimization, rendering the process unstable and forcing the policy into \textit{premature stagnation} or \textit{reward hacking}.
Existing strategies, such as varying the initial noise or increasing group sizes, often fail to address this fundamental issue, resulting in \textit{training instability} or \textit{diminishing returns}. 
To overcome these challenges, we propose \textbf{\ours (\ourshort)}, a novel framework that sustains optimization through embedding-level perturbation. 
Our method introduces structured, embedding-level perturbations within sample groups, guaranteeing a robust variance that preserves the discriminative signal throughout the training process. Extensive experiments demonstrate that our approach significantly outperforms state-of-the-art baselines, achieving a more faithful alignment with human preference.

\end{abstract}    
\section{Introduction}
\label{sec:intro}

Driven by the success of Reinforcement Learning in aligning Large Language Models with human intent~\cite{deepseek, llmalign}, RL has recently emerged as a pivotal approach for aligning diffusion and flow models with human preferences. 
While these generative models have achieved remarkable high-fidelity results~\cite{diffusion,diffusion2,chen2026stochastic,zhu2024instantswap,fang2025integrating,zhu2026artifact,ma2024followpose,ma2025followyourclick,ma2024followyouremoji,rfsolver,wang2026precisecache,wang2026elastic,mao2026omni,wu2026imagerysearch,huang2025mate,fang2026prism,chen2026conceptweaver,genreaware,yang2026macedance,feng2025narrlv,ma2025followfaster,ma2025followyourmotion,ma2025followcreation,long2025follow,shen2025follow,ma2025controllable}, large-scale pretraining alone often fails to meet specific aesthetic or semantic expectations. Consequently, RL-based optimization has been widely adopted to directly fine-tune these models against reward signals, bridging the gap between raw generation capability and human-desired outcomes~\cite{rlhf1,rlhf2}.
To this end, recent works~\cite{flowgrpo, dancegrpo, tempflowgrpo, branchgrpo, mixgrpo,zhang2025maskfocusfocusingpolicyoptimization,ma2025stagestablegeneralizablegrpo,ma2026margrpostabilizedgrpoardiffusion} have adapted the Group Relative Policy Optimization (GRPO)~\cite{deepseek} paradigm to the visual domain. By reformulating the deterministic Ordinary Differential Equation of flow matching into a Stochastic Differential Equation (SDE), these methods enable exploration and optimization directly during the sampling process. Furthermore, \cite{diffusionnft} attempts to optimize within the forward process, further improving optimization efficiency.

However, despite these advancements, a critical limitation persists: the unconstrained nature of these frameworks often erodes intra-group discriminative variance, preventing the model from distinguishing individual samples\cite{xiao2024survey, he2023strategic, he2023camouflaged, he2025segment}. 
In Fig.~\ref{fig:teaser}, this deficiency leads to two fundamental challenges:
\textbf{Challenge 1: Vanishing Discriminative Signal and Optimization Stagnation.}
Most existing methods rely excessively on relative advantage estimation, a process inherently dependent on the variance within each group. 
As training progresses, the policy naturally converges towards a narrow mode, rendering the intra-group discriminative variance negligible. This loss of variance causes the guidance signal to vanish or become numerically unstable, directly leading to premature optimization stagnation.
\textbf{Challenge 2: Restricted Exploration and Susceptibility to Reward Hacking.}
The model's search for optimal solutions becomes narrow. 
Lacking a mechanism to encourage inter-group distinctiveness, the policy converges on the nearest local optima that yield high rewards, even if these are shallow. 
This confinement to a few high-reward modes is a classic example of reward hacking~\cite{hacking}, where the model eventually collapses into a single trajectory. 
Such a collapse signifies a severe loss of generation fidelity and diversity, ultimately failing to capture the full richness of human intent.

Addressing these systemic failures often proves inadequate through existing heuristics. 
Increasing the number of parallel SDE trajectories to form a larger group is a naive approach that incurs linear computational costs while yielding diminishing returns, failing to fundamentally resolve the challenge in later training stages.
Alternatively, attempts to increase variance using distinct initial noise samples present their own set of risks.
As observed in \cite{dancegrpo}, this strategy can lead to \textit{reward hacking}, where the model exploits the reward function to generate high-scoring but visually incoherent artifacts.
To bridge this gap, we propose a novel framework, \textbf{\ours (\ourshort)}.
Our approach introduces a set of learned and structured perturbation conditions directly into the text embedding space. By systematically modulating these latent representations, \ourshort enforces a controlled intra-group discriminative variance, ensuring that each sample within a group maintains a distinct identity. This structured variance prevents the collapse of the relative discriminative signal and provides a stable, continuous gradient for fine-tuning.

In summary, our contributions are as follows:

\begin{itemize}
    \item We provide a systematic analysis of a critical failure mode in RL-based optimization: the collapse of intra-group discriminative variance. Our analysis demonstrates that this collapse directly causes optimization stagnation and reward hacking, thereby revealing a fundamental limitation of existing methods.
    
    \item We introduce \textbf{\ourshort}, a novel framework that enforces variance by injecting structured perturbations into the text embedding space. This mechanism actively guides the model to navigate varied semantic pathways, preventing it from settling into suboptimal, narrow solutions and thereby mitigating reward hacking for a more stable and faithful alignment with human intent.
    
    \item Through extensive experiments, we show that the stable optimization trajectory provided by \ourshort directly translates into superior performance. Our method simultaneously achieve high alignment scores and exceptional generation diversity, outperforming state-of-the-art baselines.
\end{itemize}


\section{Related Work}
\label{sec:related}

\subsection{RL for Visual Generation}
The application of RL to visual generation has seen significant advancements. 
Early works in this area can be broadly categorized into two primary directions. 
The first~\cite{blacktraining,rl2-dpok,policy1,diversity2,policy2,policy3} integrated RL into diffusion models by optimizing the score function through policy gradient methods. An alternative approach~\cite{rlhf1,refl,differentiable1,differentiable2,chen2026unveiling,han2026vigorbenchfarvisualgenerative,huang2026sam} focuses on direct fine-tuning with differentiable rewards. 
More recently, methods inspired by the principles of GRPO~\cite{flowgrpo,dancegrpo,prefgrpo,branchgrpo,tempflowgrpo,mixgrpo,zhang2026kvpoodenativegrpoautoregressive} have elevated the performance to a new state-of-the-art, showing strong performance and inspiring subsequent research.

\subsection{Optimization on Textual Conditions}
The quality of text-to-image generation is intrinsically linked to the input textual conditions\citep{jin2024tod3cap}. Consequently, manipulating text prompts—either in discrete token space or continuous embedding space—has become a prevalent strategy for enhancing model performance.
Some approaches~\cite{prompt1, prompt2, prompt3, prompt4,guo2026less} focused on prompt engineering, identifying specific keywords that trigger higher-quality outputs. These methods effectively align the input text with the generative model's preferred distribution, significantly improving aesthetic quality and semantic completeness. 
Beyond discrete tokens, optimization in the continuous text embedding space offers more granular control. 
\cite{textual, dreambooth} learn specific token embeddings to capture unique subjects or styles, enabling high-fidelity personalization. ~\cite{reneg, manifoldt, prompt5, chen2025contextflow, yan2025eedit, zhang2025magiccolor, feng2025dit4edit,zhao2026extending,zhou2025fireedit,liu2025controllable} optimize the unconditional branch or text embeddings to manipulate generated images accurately. These works demonstrate that subtle shifts in the text condition space can profoundly influence the generative trajectory\cite{liu2025omnidiff}.
While these existing methods demonstrate the efficacy of text condition manipulation, they primarily function as static pre-processing steps for quality enhancement\citep{shen2025uncertainty,he2024diffusion, Xiao2026Beyond, Xiao2026Quali, he2023reti, he2025unfoldldm} or reconstruction targets for personalization. In contrast, our framework uniquely integrates text perturbation directly into the online RL loop. Instead of merely refining quality\cite{he2025reversible, he2026refining}, we \textit{leverage prompt perturbation as a dynamic mechanism to induce intra-group discriminative variance}, thereby preventing optimization stagnation and mitigating reward hacking during the alignment process.

\section{Preliminaries}
\label{sec:preliminary}

\subsection{Flow Matching}
\label{subsec:flow_matching}
Let $x_0 \sim \mathcal{X}_0$ denote a data sample from the true distribution and $x_1 \sim \mathcal{X}_1$ a noise sample from the standard Gaussian $\mathcal{N}(0, I)$. Recent generative models adopt the Rectified Flow framework~\cite{flow1,flow2}, which defines the intermediate state $x_t$ at time $t \in [0, 1]$ via linear interpolation:
\begin{equation}
    x_t = (1 - t)x_0 + t x_1.
    \label{eq:rf_interp}
\end{equation}
To enable controllable generation, a flow model $v_\theta(x_t, t, c)$ is trained to regress the target $v = x_1 - x_0$ conditioned on context $c$ (e.g., class label or text embedding) by minimizing the Flow Matching objective:
\begin{equation}
    \mathcal{L}_{\text{FM}}(\theta) = \mathbb{E}_{t \sim \mathcal{U}[0,1], x_0, x_1, c} \left[ \| v_\theta(x_t, t, c) - v \|^2 \right].
    \label{eq:fm_loss}
\end{equation}


\subsection{Diffusion Negative-aware Finetuning}
\label{subsec:diffusion_nft}
Unlike policy gradient methods that require formalizing denoising as an MDP and differentiating through solvers~\cite{flowgrpo,dancegrpo,tempflowgrpo}, DiffusionNFT~\cite{diffusionnft} directly optimizes the forward process using a contrastive objective. 
A group of $K$ images $\{\bm{x}_0^{(k)}\}_{k=1}^K$ generated online from the old policy $v^{\text{old}}$ using arbitrary solvers. The optimality probability $r(x_0, c) \in [0, 1]$ is derived by normalizing the raw reward $r^{\text{raw}}$ with a stabilizing factor $Z_c$:
\begin{equation}
\label{eq:r_normalization}
\begin{split}
    r(\bm{x}_0, \bm{c}) := & \frac{1}{2} + \frac{1}{2} \operatorname{clip} \bigg[  \frac{1}{Z_{\bm{c}}} \Big( r^{\mathrm{raw}}(\bm{x}_0, \bm{c}) \\
    & - \mathbb{E}_{\pi^{\mathrm{old}}(\cdot|\bm{c})} r^{\mathrm{raw}}(\bm{x}_0, \bm{c}) \Big), -1, 1 \bigg].
\end{split}
\end{equation}

The objective is to steer the velocity field $v_\theta$ towards high-reward regions ($r \to 1$) and away from low-reward ones ($r \to 0$) via a weighted contrastive loss:
\begin{equation}
\begin{split}
    \mathcal{L}_{\text{NFT}}(\theta) = \mathbb{E}_{t, c, x_0, x_1} \big[ &r \| v_\theta^+(x_t, t, c) - v \|^2 \\
    &+ (1 - r) \| v_\theta^-(x_t, t, c) - v \|^2 \big],
\end{split}
    \label{eq:nft_loss}
\end{equation}
where $v$ is the target velocity field corresponding to the sample $x_0$. The implicit positive ($v_\theta^+$) and negative ($v_\theta^-$) policies are parameterized interpolations between the old policy $v^{\text{old}}$ and the training policy $v_\theta$, controlled by $\beta$:
\begin{align}
    v_\theta^+(x_t, t, c) &:= (1 - \beta) v^{\text{old}}(x_t, t, c) + \beta v_\theta(x_t, t, c), \label{eq:implicit_pos} \\
    v_\theta^-(x_t, t, c) &:= (1 + \beta) v^{\text{old}}(x_t, t, c) - \beta v_\theta(x_t, t, c). \label{eq:implicit_neg}
\end{align}
To ensure training stability, the standard velocity matching loss is replaced by a self-normalized $\bm{x}_0$-regression:
\begin{equation}
\label{eq:nft_loss_x0}
\begin{aligned}
    \mathcal{L}_{\text{NFT}}(\theta) = \mathbb{E}_{t, c, x_0, x_1} \bigg[  r \cdot \frac{\| \bm{x}_{\bm{v}_\theta^+}(x_t, t, c) - \bm{x}_0 \|_2^2}{\operatorname{sg}(\mu(|\bm{x}_{\bm{v}_\theta^+}(x_t, t, c) - \bm{x}_0|))} \\ + (1 - r) \cdot \frac{\| \bm{x}_{\bm{v}_\theta^-}(x_t, t, c) - \bm{x}_0 \|_2^2}{\operatorname{sg}(\mu(|\bm{x}_{\bm{v}_\theta^-}(x_t, t, c) - \bm{x}_0|))} \bigg],
\end{aligned}
\end{equation}
where $\mu(\cdot)$ denotes the mean operator, $\operatorname{sg}(\cdot)$ is the stop-gradient, $\bm{x}_{\bm{v}_\theta^+}$ and $\bm{x}_{\bm{v}_\theta^-}$ represent the data predicted by the $v_\theta^+$ and $v_\theta^-$.
This formulation allows for efficient likelihood-free training and enables training and inference without Classifier-Free Guidance~\cite{cfg}, internalizing the guidance direction by negative-aware finetuning.
It naturally incorporates reinforcement signals into the supervised learning objective, effectively unifying online optimization with the standard flow matching paradigm.
\label{subsec:variance_via_condition}

\section{Methodology}
\label{sec:method}
\begin{figure*}[!t] 
    \centering
    \captionsetup[subfigure]{font=scriptsize}
    \begin{subfigure}[b]{0.33\textwidth} 
        \centering
        \includegraphics[width=\linewidth]{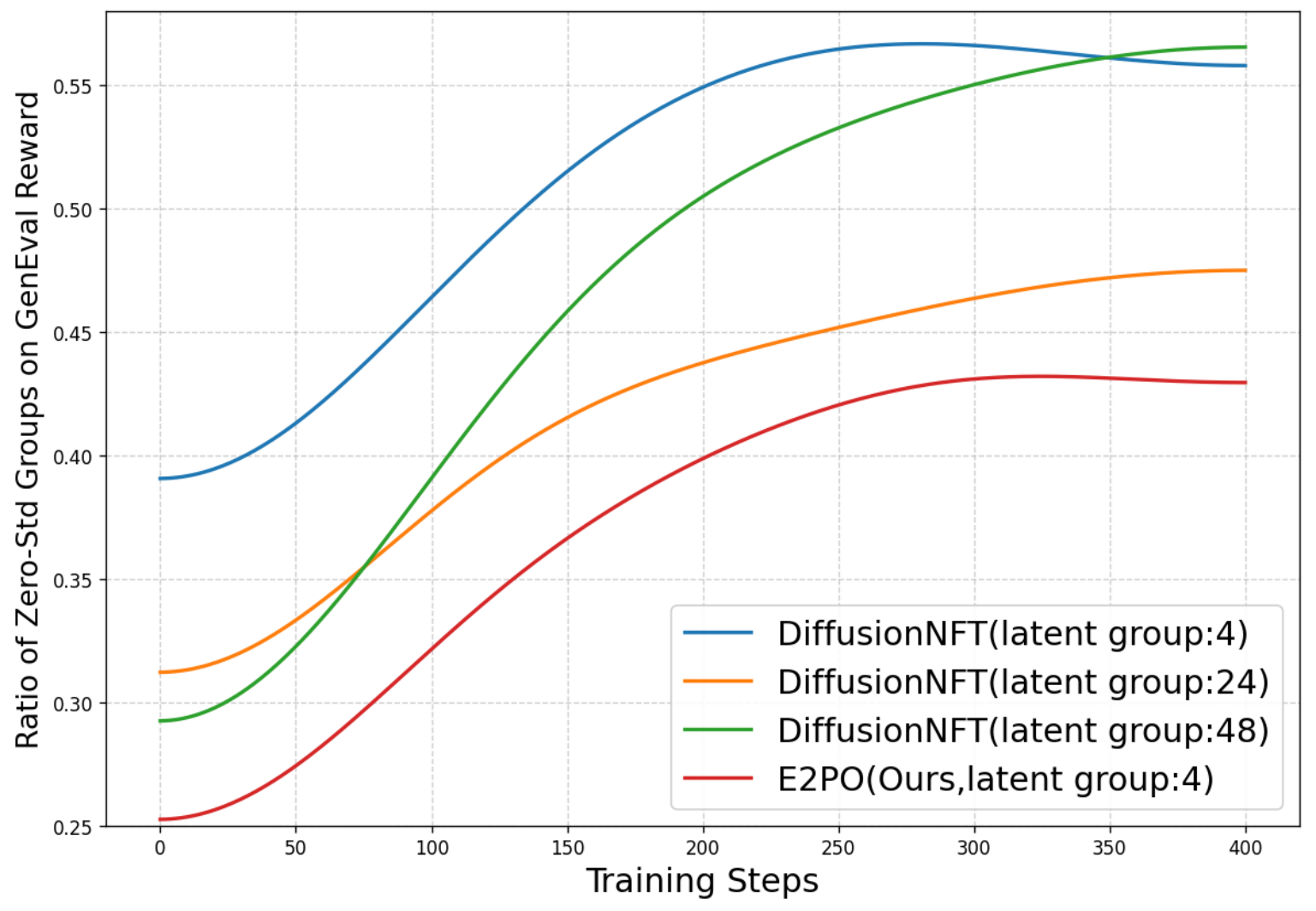} 
        \caption{Ratio of groups with zero standard deviation}
        \label{fig:std_a}
    \end{subfigure}
    \hfill 
    \begin{subfigure}[b]{0.33\textwidth}
        \centering
        \includegraphics[width=\linewidth]{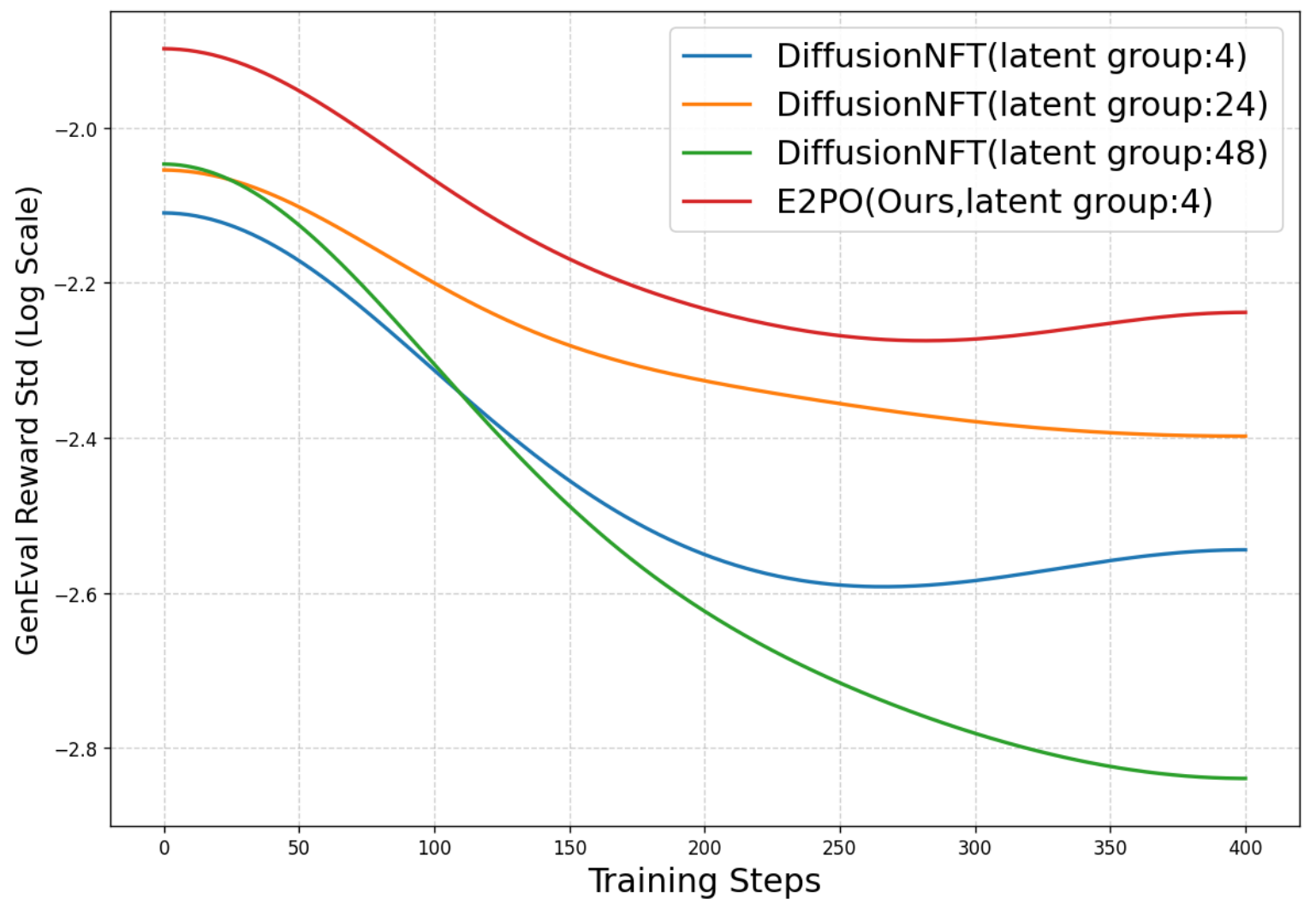}
        \caption{GenEval reward standard deviation (log scale)}
        \label{fig:std_b}
    \end{subfigure}
    \hfill 
    \begin{subfigure}[b]{0.33\textwidth}
        \centering
        \includegraphics[width=\linewidth]{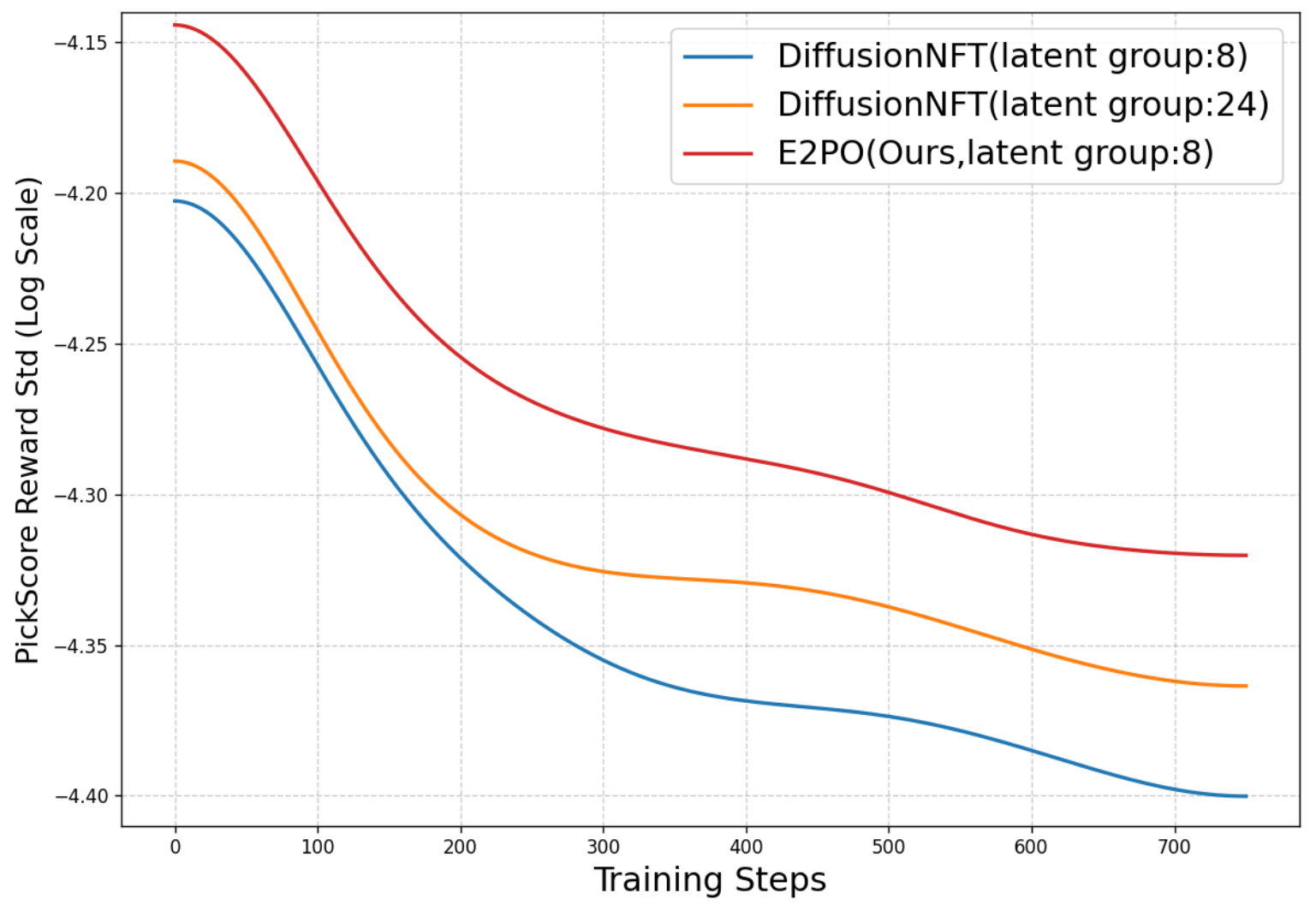}
        \caption{PickScore reward standard deviation(log scale)}
        \label{fig:std_c}
    \end{subfigure}
    \caption{\textbf{Evolution of Intra-Group Discriminative Variance During Training.} Smoothed curves track the changing trends of variance statistics as training proceeds. The baseline's standard deviation decline significantly, whereas \ourshort maintains a consistent variance level, demonstrating sufficient intra-group discriminative variance throughout the optimization process.}
    \vspace{-10pt}
    \label{fig:std}
\end{figure*}
In this section, we first analyze the role of intra-group discriminative variance and the challenges posed by its decay. 
Building on this analysis, we introduce \ours (\ourshort), which addresses these challenges by perturbing the conditional embedding. 
This approach ensures optimization stability by sustaining variance and expands the exploration space to effectively mitigate reward hacking.

\subsection{Intra-group Discriminative Variance}
Effective optimization hinges on maintaining sufficient intra-group discriminative variance. 
However, as training progresses, this variance often collapses ($\sigma_R \to 0$), which nullifies the optimization signal and causes the learning process to stagnate. 
The root of this problem lies in how the gradient is calculated. 
As formulated in Eq.~\ref{eq:r_normalization} and Eq.~\ref{eq:nft_loss}, where denotes the optimality probability~$r$, which steers the gradient, is normalized by the group reward standard deviation~$\sigma_R$. 
As illustrated by the example of discrete rewards in Fig.~\ref{fig:std_a}, this dependency means that a near-zero~$\sigma_R$ cripples the learning signal, resulting in \textbf{Vanishing Discriminative Signal and Optimization Stagnation (Challenge 1)}.
To mitigate variance collapse, current methods, which rely on stochastic noise for exploration, often increase the group size to maintain discriminative variance. However, this strategy yields diminishing returns: as shown in Fig.~\ref{fig:std_a} and~\ref{fig:std_b}, increasing the group size~$G$ from 4 to 48 incurs substantially higher computational costs yet fails to arrest the decay.
This failure prompts a critical question: \textit{can we find a more effective source of diversity by looking beyond the noise prior}?

Drawing inspiration from recent advancements in text condition manipulation (discussed in Sec.~\ref{sec:related}), we observe that the text embedding space offers a structured alternative to unstructured stochastic noise. Within this space, even subtle shifts can induce meaningful semantic variations while preserving alignment with the core intent. We posit that this mechanism offers a potent solution to the variance collapse problem, \textit{enabling the discovery of diverse sampling trajectories that are otherwise inaccessible via standard latent noise}.
To validate this hypothesis, we adopt a strategy that shifts the locus of exploration. Specifically, instead of relying on computationally expensive latent scaling (e.g., increasing $G$ from 4 to 48), we maintain a minimal latent group size (e.g., $G=4$) and actively inject $K$ semantic perturbations for each seed. Formally, we redefine the exploration set as:
\vspace{-5pt}
\begin{equation}
    \{ \Phi_\theta(x_1^{(i)}, c) \}_{i=1}^{G_1} \xrightarrow{\text{Ours}} \{ \Phi_\theta(x_1^{(j)}, c + \delta_k) \}_{j=1, k=1}^{G_2, K},
\end{equation}

where $G_1=G_2 \times K$. 
By explicitly modulating the condition, we empirically validate our hypothesis: as shown in Fig.~\ref{fig:std_b} and~\ref{fig:std_c}, this approach sustains significantly higher reward variance and maintains a lower zero-std ratio (Fig.~\ref{fig:std_a}) compared to the brute-force latent baseline. This confirms that active perturbation in the embedding space induces a more robust optimization signal, effectively preventing premature stagnation.

Having empirically validated the potential of embedding perturbation to sustain optimization variance, we now address the deeper issue of \textbf{Restricted Exploration and Susceptibility to Reward Hacking (Challenge 2)}.

\begin{figure*}[t] 
    \centering
    \includegraphics[width=0.95\textwidth]{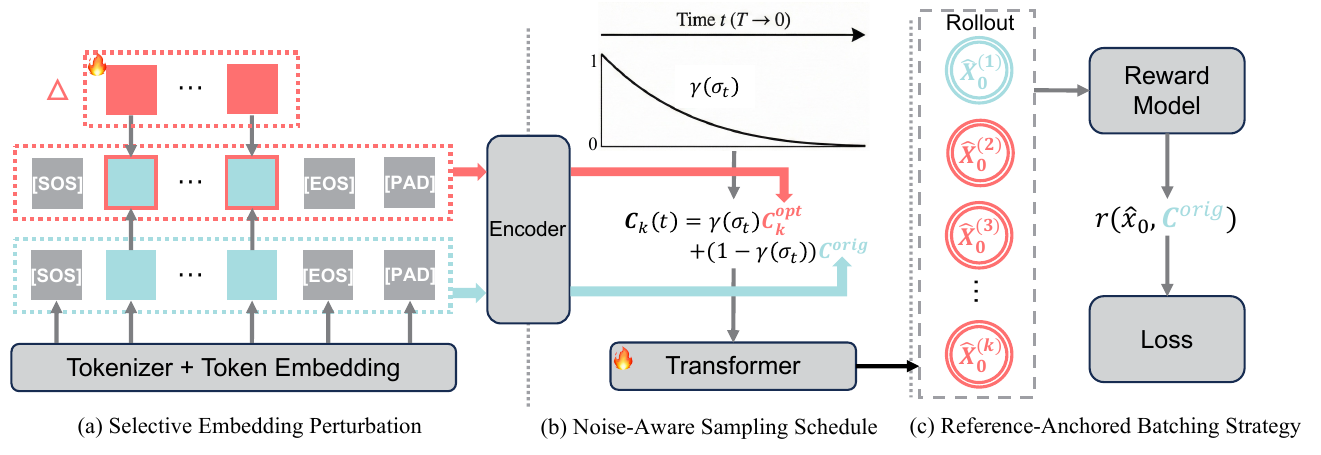} 
    \caption{\textbf{Method Overview.} 
We introduce (a) selective perturbation on content embeddings to induce discriminative signal, 
(b) a noise-aware schedule to modulate condition injection during sampling, and 
(c) a reference-anchored strategy that calculates gradients relative to the original prompt $C^{\text{orig}}$ to prevent semantic drift.}
    \label{fig:pipeline}
    \vspace{-10pt} 
\end{figure*}

\subsection{\ours}
Robust RL necessitates varied sample coverage to prevent the policy from overfitting to narrow high-reward regions or exploiting proxy shortcuts~\cite{hacking, chen2025taming,liu2025diversegrpo,wang2025grpo}. However, as training progresses, generative models naturally gravitate toward probability-dense \textit{``strong modes"}~\cite{manifoldc, manifoldt, manifoldf}. This collapse into homogeneous outputs severely restricts the exploration space, making the policy susceptible to \textit{reward hacking} by exploiting fixed, non-semantic artifacts.
To effectively expand the exploration boundary beyond these collapsed modes, we turn to the intrinsic geometry of the semantic space. According to \cite{manifoldf, manifoldb, manifoldc}, valid semantic variations do not span the entire high-dimensional space uniformly but instead concentrate near low-dimensional structures. 
While standard stochastic noise fails to reach these meaningful variations, navigating this semantic manifold allows the policy to uncover ``underrepresented regions''—trajectories that are semantically distinct from the dominant mode yet remain plausible.

Leveraging this insight, we propose \textbf{\ours (\ourshort)}. 
Specifically, our approach injects perturbations within the embedding space at every training step, forcing the policy to constantly navigate diverse semantic variations throughout the entire optimization process.
This persistent perturbation prevents the model from stagnating in narrow, high-reward ``traps." The efficacy of this strategy is visualized in the right inset of Fig.~\ref{fig:teaser}, which projects the high-dimensional DINOv3 features of sampled images into a 2D plane using t-SNE. While the baseline suffers from restricted exploration leading to vanishing discriminative signal, our approach facilitates broader exploration. Notably, even at 300 steps, our method maintains significantly superior semantic coverage compared to the baseline.

To systematically implement this framework, \ourshort incorporates three key components as illustrated in Fig.~\ref{fig:pipeline}: (1) an \textit{Embedding-Perturbed Mechanism} to generate diverse semantic variants, (2) a \textit{Noise-Aware Sampling Schedule} to steer generation trajectories, and (3) a \textit{Reference-Anchored Batching Strategy} to stabilize policy updates, detailed algorithm in \supp~\ref{appendix:algorithm}

\begin{figure*}[t] 
    \centering
    \includegraphics[width=0.95\linewidth]{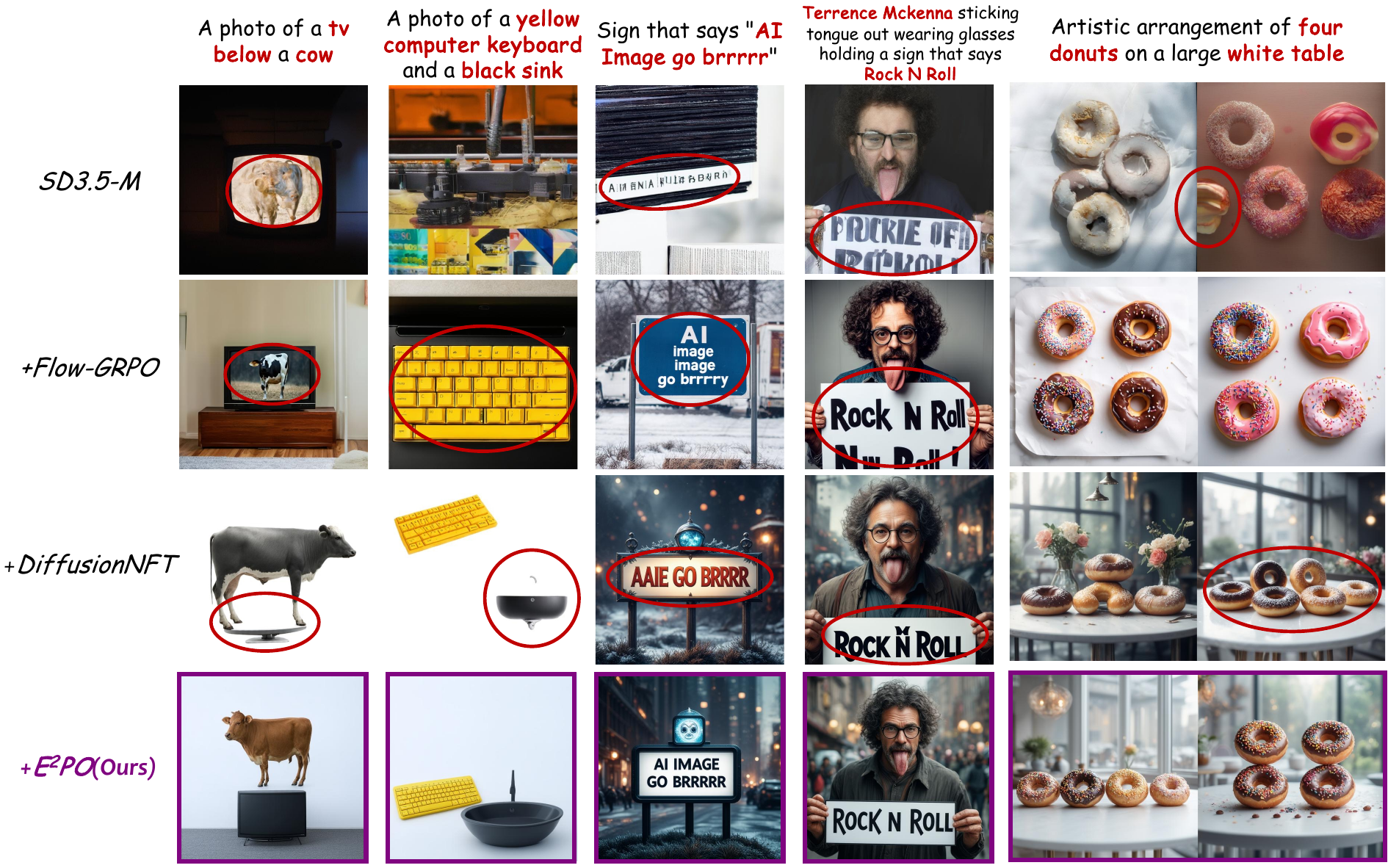}
    \caption{\textbf{Qualitative Comparison of \ourshort against Baselines.} \ourshort demonstrates superior performance fidelity, spatial reasoning, instruction adherence and diversity, overcoming the limitations (highlighted in red circles) seen in other methods.}
    \label{fig:visualization_merge}
    \vspace{-10pt} 
\end{figure*}

\textbf{Embedding-Perturbed Mechanism.} 
To achieve this, we devise an embedding-perturbed mechanism. Specifically, given a prompt $p$, we first obtain the tokenized sequence $\mathbf{x} \in \mathbb{Z}^S$, which is then mapped to the static embeddings $\mathbf{E} \in \mathbb{R}^{S \times d}$. To restrict exploration to effective semantic content, we define an optimizing index set $\mathcal{I}$ that excludes padding and syntactic boundary tokens (i.e., \texttt{[SOS]}, \texttt{[EOS]}), yielding an effective content length $L = |\mathcal{I}|$.

We parameterize the active exploration via a set of $K$ \textit{learnable} perturbation tensors $\Delta = \{\bm{\delta}_k\}_{k=1}^K$, where each $\bm{\delta}_k \in \mathbb{R}^{L \times d}$ is initialized from a Gaussian prior:
\begin{equation}
    \bm{\delta}_k \sim \mathcal{N}(0, \sigma_{\text{init}}^2 \mathbf{I}),
\end{equation}
where $\sigma_{\text{init}}$ governs the initialization magnitude. To ensure these perturbations translate into \textit{valid semantic shifts}, we selectively target effective content tokens while freezing others. The perturbed input representation $\tilde{\mathbf{E}}_k \in \mathbb{R}^{S \times d}$ is constructed via:
\begin{equation}
    \label{eq:embedding_perturbed}
    \tilde{\mathbf{E}}_{k, t} =
    \begin{cases}
        \mathbf{E}_t + \bm{\delta}_{k, j} & \text{if } t = \mathcal{I}_j \\
        \mathbf{E}_t & \text{otherwise}
    \end{cases},
\end{equation}
where $t$ indexes the token position and $\mathcal{I}_j$ denotes the coordinate of the $j$-th content token. This sequence is then processed by the frozen text encoder $f_\phi(\cdot)$ to compute the contextualized features.
Following initialization, we optimize these perturbations to achieve \textit{an optimal trade-off between exploration diversity and semantic consistency}. Let $e_k = \operatorname{Pool}(f_\phi(\tilde{\mathbf{E}}_k))$ and $e_{\text{anc}} = \operatorname{Pool}(f_\phi(\mathbf{E}))$ denote the global semantic embeddings for the $k$-th variant and the unperturbed anchor, respectively. We construct a hybrid objective to optimize $\Delta$. First, to promote semantic diversity and prevent redundancy, we minimize the mean pairwise cosine similarity among the variants:
\begin{equation}
    \label{eq:diversity}
    \mathcal{L}_{\text{div}}(\Delta) = \frac{1}{K(K-1)} \sum_{i=1}^K \sum_{j \neq i} \frac{e_i^\top e_j}{\|e_i\| \|e_j\|}.
\end{equation}
Simultaneously, to strictly confine exploration within a valid semantic neighborhood, we impose a geometric constraint that anchors variants around a target similarity $\mu$ relative to the original prompt:
\begin{equation}
    \label{eq:similariyu}
    \mathcal{L}_{\text{anc}}(\Delta) = \sum_{k=1}^K \left| \frac{e_k^\top e_{\text{anc}}}{\|e_k\| \|e_{\text{anc}}\|} - \mu \right|_\epsilon,
\end{equation}
where $|\cdot|_\epsilon$ denotes the $\epsilon$-insensitive loss. The final objective is a weighted sum: $\mathcal{L}_{\text{emb}} = \lambda_{div} \mathcal{L}_{\text{div}} + \mathcal{L}_{\text{anc}}$, where $\lambda_{div}$ governs the trade-off between semantic diversity and consistency.

Upon completing the embedding optimization, we freeze the learned perturbations and proceed to policy training. 
Let $\mathbf{C}^{\text{orig}}$ and $\mathbf{C}^{\text{opt}}_k$ denote the conditioning contexts derived from the original and the $k$-th optimized embeddings, respectively.

\textbf{Noise-Aware Sampling Schedule.} 
Recognizing that the diffusion model's sensitivity varies across timesteps, we implement a noise-aware sampling schedule, dynamically interpolating the condition $\mathbf{C}_k(t)$ based on the normalized sampling time $\sigma_t$ (decaying from $1 \to 0$):
\begin{equation}
    \label{eq:noise_aware}
    \mathbf{C}_k(t) = \gamma(\sigma_t) \mathbf{C}^{\text{opt}}_k + (1 - \gamma(\sigma_t)) \mathbf{C}^{\text{orig}},
\end{equation}
where $\gamma(\sigma_t) = \operatorname{clip}\left( \frac{\sigma_t - (1 - \rho)}{\rho}, 0, 1 \right)$ is controlled by a perturbation range $\rho \in (0, 1]$. This allows semantic variants to steer the trajectory toward diverse structural layouts during the early high-noise phase, while smoothly reverting control to the unperturbed anchor in the later stages to ensure fine-grained visual fidelity.

\textbf{Reference-Anchored Batching Strategy.} 
Complementing the trajectory steering, we employ a reference-anchored batching strategy to construct the final exploration set. Specifically, we select $K-1$ optimized conditions and explicitly retain the original condition as a stable anchor: $\mathcal{C}_{\text{batch}} = \{ \mathbf{C}^{\text{orig}} \} \cup \{ \mathbf{C}_k(t) \}_{k=1}^{K-1}$. Sampling from this composite $\mathcal{C}_{\text{batch}}$ yields a rich set of trajectories, providing a diverse basis for the subsequent policy update.
Finally, we \textit{integrate the discovered trajectories into the policy optimization process}. 
Crucially, aligning with the supervised RL paradigm in Sec.~\ref{subsec:diffusion_nft}, we decouple exploration from optimization: samples are generated using the perturbed guidance $\mathbf{C}_k(t)$, whereas policy updates are calculated by conditioning strictly on the unperturbed anchor $\mathbf{C}^{\text{orig}}$.
By optimizing with this objective, the model effectively steers the distribution of the original prompt toward high-reward structural regions while penalizing trajectories associated with low-reward outcomes.

\section{Experiments}
\label{sec:exp}

\begin{table*}[t]
\centering
\caption{\textbf{Performance Comparison of RL Methods on GenEval.} All competing methods are trained using GenEval as the reward model across varying latent group sizes $G$. For \ourshort, we employ two distinct experimental settings: \textbf{High-Exploration} ($K=12$) and \textbf{Efficient Generation} ($K=4$). \ourshort consistently achieves superior performance on in-domain reward and diversity. \colorbox{graybg}{Gray background}: In-domain reward. \textbf{Bold}: Best. \underline{Underline}: Second best. \textit{Note: Flow-GRPO exceeds memory limits at $G=48$.}}
\label{tab:geneval}
\resizebox{\textwidth}{!}{%
\begin{tabular}{l c c ccccc c c ccccc}
\toprule
\multirow{3}{*}{Method} & & \multicolumn{6}{c}{\textbf{Number of Semantic Variants $K=12$}} & & \multicolumn{6}{c}{\textbf{Number of Semantic Variants $K=4$}} \\
\cmidrule(lr){3-8} \cmidrule(lr){10-15}
& & $G$ & Reward & \multicolumn{4}{c}{Diversity} & & $G$ & Reward & \multicolumn{4}{c}{Diversity} \\
\cmidrule(lr){4-4} \cmidrule(lr){5-8} \cmidrule(lr){11-11} \cmidrule(lr){12-15}
& & & GenEval $\uparrow$ & IDS $\downarrow$ & ASC $\uparrow$ & SDI $\uparrow$ & PVS $\uparrow$ & & & GenEval $\uparrow$ & IDS $\downarrow$ & ASC $\uparrow$ & SDI $\uparrow$ & PVS $\uparrow$ \\
\midrule
SD3.5-M & & {\color{gray} ---} & {\color{gray} 0.263} & {\color{gray} 0.044} & {\color{gray} 0.143} & {\color{gray} 0.458} & {\color{gray} 0.392} & & {\color{gray} ---} & {\color{gray} 0.263} & {\color{gray} 0.044} & {\color{gray} 0.143} & {\color{gray} 0.458} & {\color{gray} 0.392}  \\
\midrule

\multirow{2}{*}{+Flow-GRPO}    
& & 4  & \cellcolor{graybg}0.724 & \underline{0.051} & \underline{0.126} & 0.462 & \underline{0.321} & & 2 & \cellcolor{graybg}0.678 & 0.079 & 0.118 & \underline{0.452} & \textbf{0.195} \\
& & 24 & \cellcolor{graybg}0.776 & 0.064 & 0.123 & 0.422 & 0.318 & & 8 & \cellcolor{graybg}0.769 & 0.081 & 0.125 & 0.447 & 0.190 \\
\midrule

\multirow{3}{*}{+DiffusionNFT} 
& & 4  & \cellcolor{graybg}0.868 & 0.077 & \underline{0.126} & 0.458 & 0.260 & & 2 & \cellcolor{graybg}0.822 & \underline{0.076} & 0.042 & 0.450 & \underline{0.194} \\
& & 24 & \cellcolor{graybg}\underline{0.922} & 0.054 & 0.118 & 0.418 & 0.259 & & 8 & \cellcolor{graybg}\underline{0.915} & 0.093 & \underline{0.128} & 0.417 & 0.188 \\
& & 48 & \cellcolor{graybg}0.917 & \underline{0.051} & 0.109 & \underline{0.463} & 0.196 & & --- & --- & --- & --- & --- & --- \\
\midrule

\textbf{+\ourshort (Ours)} 
& & 4  & \cellcolor{graybg}\textbf{0.932} & \textbf{0.048} & \textbf{0.127} & \textbf{0.467} & \textbf{0.322} & & 2 & \cellcolor{graybg}\textbf{0.917} & \textbf{0.075} & \textbf{0.136} & \textbf{0.453} & \textbf{0.195} \\

\bottomrule
\end{tabular}
}
\end{table*}

\begin{table*}[t]
\centering
\caption{\textbf{Performance Comparison of RL Methods on PickScore.} All competing methods are trained using PickScore as the reward model across varying latent group sizes $G$. For \ourshort, we employ a fixed number of semantic variants ($K=3$). \ourshort demonstrates superior alignment with human preference metrics while maintaining diversity. \colorbox{graybg}{Gray background}: In-domain reward. \textbf{Bold}: Best. \underline{Underline}: Second best. }
\label{tab:pickscore}
\small
\resizebox{0.9\textwidth}{!}{%
\begin{tabular}{ll c c c cc cccc} 
\toprule
\multirow{2}{*}{Method} & & \multirow{2}{*}{$G$} & \multicolumn{1}{c}{Reward} & \multicolumn{1}{c}{Quality} & \multicolumn{2}{c}{Preference Score} & \multicolumn{4}{c}{Diversity} \\
\cmidrule(lr){4-4} \cmidrule(lr){5-5} \cmidrule(lr){6-7} \cmidrule(lr){8-11}
 & & & PickScore $\uparrow$ & Aesthetic $\uparrow$ & ImgRwd $\uparrow$ & HPSv2.1 $\uparrow$ & IDS $\downarrow$ & ASC $\uparrow$ & SDI $\uparrow$ & PVS $\uparrow$ \\
\midrule
\textcolor{gray}{SD3.5-M} & & \textcolor{gray}{---} & \textcolor{gray}{19.93} & \textcolor{gray}{5.600} & \textcolor{gray}{-0.50} & \textcolor{gray}{0.203} & \textcolor{gray}{0.044} & \textcolor{gray}{0.143} & \textcolor{gray}{0.458} & \textcolor{gray}{0.392} \\
\midrule

\multirow{2}{*}{+Flow-GRPO}    
& & 8  & \cellcolor{graybg}22.60 & 6.223 & 1.27 & 0.315 & 0.220 & \underline{0.137} & \underline{0.411} & 0.280 \\
& & 24 & \cellcolor{graybg}22.72 & 6.273 & \textbf{1.30} & \underline{0.324} & 0.222 & 0.131 & 0.410 & 0.276 \\
\addlinespace[0.5em]

\multirow{2}{*}{+DiffusionNFT} 
& & 8  & \cellcolor{graybg}23.31 & \underline{6.535} & 1.25 & 0.323 & \underline{0.217} & \underline{0.137} & 0.409 & 0.282 \\
& & 24 & \cellcolor{graybg}\underline{23.34} & 6.514 & 1.27 & \underline{0.324} & 0.239 & 0.133 & 0.390 & \underline{0.298} \\
\addlinespace[0.5em]

\textbf{+\ourshort(Ours)}          
& & 8  & \cellcolor{graybg}\textbf{23.38} & \textbf{6.538} & \underline{1.29} & \textbf{0.325} & \textbf{0.167} & \textbf{0.138} & \textbf{0.412} & \textbf{0.301} \\

\bottomrule
\end{tabular}
}
\vspace{-10pt}
\end{table*}

\subsection{Implementation Details}
\textbf{Experimental Setup.} 
We utilize Stable Diffusion 3.5 Medium (SD3.5-M)~\cite{diffusion2} as the backbone, conducting all experiments on 8$\times$ H20 GPUs. 
Defining the exploration space via latent group size $G$ (noise seeds) and semantic variants $K$ (embedding perturbations), we evaluate \ourshort across three configurations: 
(i) \textbf{High-Exploration Compositional Generation} ($G=4, K=12$) on GenEval to assess peak reasoning; 
(ii) \textbf{Efficient Compositional Generation} ($G=2, K=4$) to demonstrate sample efficiency; and 
(iii) \textbf{Human Preference Alignment} ($G=8, K=3$) on PickScore for visual fidelity. 
Performance is assessed using direct task metrics (GenEval~\cite{ghosh2023geneval}, PickScore~\cite{pickscore}), generalization to unseen rewards (e.g., Aesthetic Score~\cite{aes}, HPSv2.1~\cite{hps}), and diversity benchmark (DivGenBench~\cite{chen2025taming}). Detailed training configurations and hyperparameters are provided in the \supp~\ref{appendix:exp_details}.

\subsection{Qualitative Evaluation}
The qualitative comparisons on compositional generation and human preference alignment are presented in Fig.~\ref{fig:visualization_merge}. Our method consistently outperforms the baselines in terms of fidelity, spatial reasoning, instruction adherence and diversity. 
Specifically, baselines struggle with spatial reasoning (Col. 1) and attribute binding (Col. 2), often hallucinating object positions (e.g., placing the cow \textit{inside} the TV) or failing to generate all concepts like the ``black sink" whereas \ourshort accurately synthesizes all requested attributes. Our method also excels in fine-grained text generation (Cols. 3-4), rendering precise strings with high visual fidelity where baselines produce garbled characters. Finally, \ourshort mitigates reward hacking evident in counting tasks (Col. 5); while baselines collapse into repetitive layouts or over-generate items, our approach maintains diverse structural layouts while strictly adhering to numerical constraints. More visualization results provided in the \supp~\ref{appendix:visualization}.

\subsection{Quantitative Evaluation}



As shown in Tab.~\ref{tab:geneval}, regarding GenEval, \ourshort achieves superior scores in both \textit{High-Exploration} ($K=12$) and \textit{Efficient} ($K=4$) settings, demonstrating effective search space navigation for peak reasoning and high sample efficiency across all group sizes. Regarding PickScore (Tab.~\ref{tab:pickscore}), our method attains the highest in-domain reward and strong transferability to unseen metrics. Notably, unlike baselines exhibiting lower distinctiveness, \ourshort achieves the best performance across all metrics on DivGenBench, which confirm the mitigation of over-optimization and reward hacking. Furthermore, a user study following~\cite{dancegrpo} confirms that \ourshort achieves the best win rate in human preference alignment (see \supp~\ref{appendix:user_study}).

\begin{figure*}[!t] 
    \centering
    \includegraphics[width=0.95\textwidth]{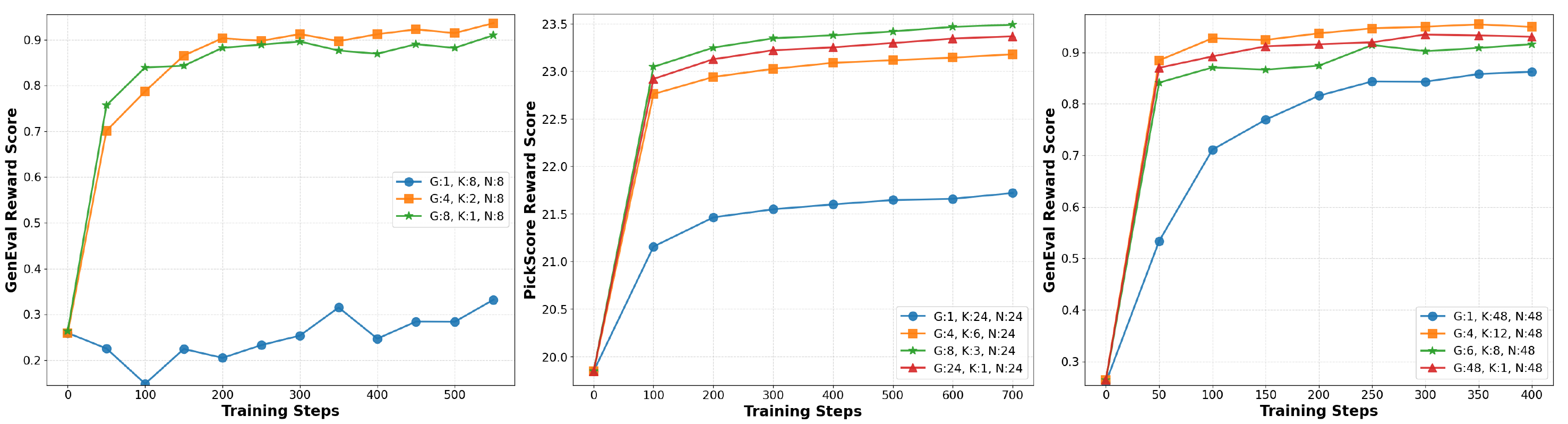} 
    \caption{\textbf{Ablation of Latent Group Size $G$ and Number of Semantic Variants $K$.} We analyze the trade-off between $G$ and $K$ under a fixed computational budget ($N = G \times K$). We observe that the extreme configurations ($G=1$ or $K=1$) are insufficient, whereas a balanced split between $G$ and $K$ achieves the most stable and highest performance.}
    \label{fig:ablation_g_k}
    \vspace{-10pt} 
\end{figure*}

\subsection{Ablation Study}

\begin{figure}[t] 
    \centering
    \includegraphics[width=\linewidth]{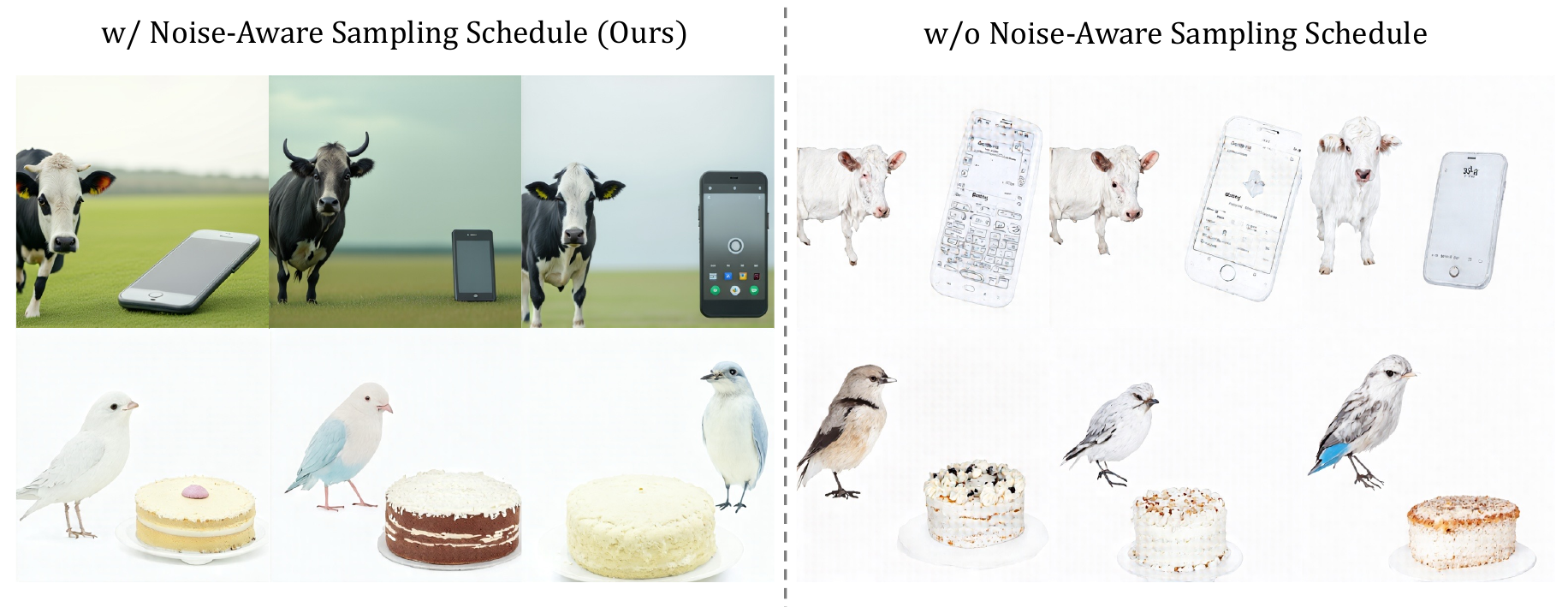} 
    \caption{\textbf{Ablation of Noise-Aware Sampling Schedule.} Samples are generated at the 150-th training step. The static strategy (right) leads to semantic drift or artifacts, whereas our Noise-Aware Schedule (left) maintains high visual fidelity.}
    \label{fig:ablation_noise_aware}
    \vspace{-10pt}
\end{figure}

\textbf{Balancing Latent Group Size and Number of Semantic Variants.} 
We analyze the trade-off between Latent Group Size $G$ and the Number of Semantic Variants $K$ under a fixed budget $N = G \times K$ (Fig.~\ref{fig:ablation_g_k}).
First, the $G=1$ configuration (single noise prior) performs poorly across all tasks, confirming that semantic perturbations alone provide insufficient exploration. This underscores the necessity of combining both noise-level and semantic-level diversity to effectively expand the exploration space. 
Second, results show that a balanced configuration achieves the best and most stable performance. In the high-budget regime ($N=48$), the setting $G=4, K=12$ significantly outperforms extreme cases where either $G=1$ or $K=1$. This proves that a suitable split between $G$ and $K$ ensures a better exploration range, leading to the higher reward.

\paragraph{Effectiveness of Noise-Aware Sampling Schedule.}
To validate our dynamic interpolation, we compare it against a static strategy where optimized variants $\mathbf{C}^{\text{opt}}_k$ are applied throughout sampling. As shown in Fig.~\ref{fig:ablation_noise_aware}, the static approach often compromises quality, causing artifacts or drift. In contrast, by restricting semantic intervention to the high-noise phase and reverting to the unperturbed anchor $\mathbf{C}^{\text{orig}}$ later, our method achieves diverse layouts while maintaining high visual quality.

\begin{figure}[!] %
    \centering
    \includegraphics[width=\linewidth]{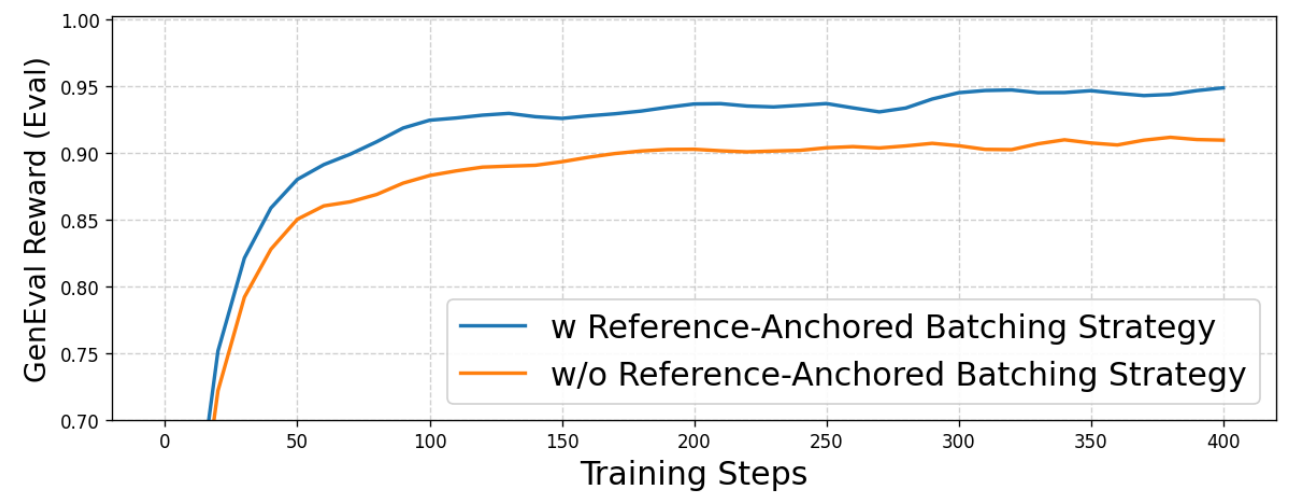} 
    \caption{\textbf{Ablation of Reference-Anchored Batching Strategy.} The model with Reference-Anchored Batching strategy exhibits a distinct performance advantage.}
    \label{fig:ablation_ref}
    \vspace{-10pt} 
\end{figure}

\textbf{Effectiveness of Reference Anchor Embeddings.} 
To assess the necessity of semantic regularization, we ablate the reference anchor embeddings $\mathbf{C}^{\text{orig}}$. As illustrated in Fig.~\ref{fig:ablation_ref}, incorporating $\mathbf{C}^{\text{orig}}$ yields consistently superior convergence and asymptotic performance compared to the unconstrained baseline. 
Due to space limitations, we present a more comprehensive investigation regarding the choice of text encoders and repulsive coefficient $\lambda_{div}$ in \supp~\ref{appendix:ablation}.
\section{Conclusion}
\label{sec:conclusion}

In this work, we address a critical failure mode in group-based RL alignment: the premature optimization stagnation caused by the decay of intra-group variance. We identify that this decay eliminates the discriminative signal essential for stable training. To counteract this, we introduce \textbf{\ours (\ourshort)}, a novel framework that maintains this signal through \textbf{structured, embedding-level} perturbations. 
Experiments show \ourshort achieves a more stable and effective alignment than state-of-the-art baselines.

\section*{Impact Statement}


This paper presents work whose goal is to advance the field of Machine
Learning. There are many potential societal consequences of our work, none
which we feel must be specifically highlighted here.


\nocite{langley00}

\bibliography{example_paper}
\bibliographystyle{icml2026}



\newpage
\appendix
\onecolumn
\section{\ourshort Algorithm}
\label{appendix:algorithm}

We summarize the complete training procedure of \ourshort in Algorithm~\ref{alg:ourshort}. 
For each sampled prompt, we use \textit{Embedding-Perturbed Mechanism} to generate diverse semantic variants, by initializing a set of learnable perturbation tensors $\Delta$ and optimizing them to maximize the semantic diversity of the conditions while maintaining consistency with the anchor, minimizing the combined objective $\mathcal{L}_{\text{emb}}$.
Subsequently, in the sampling phase, we generate trajectories using the \textit{Noise-Aware Sampling Schedule}, where the conditioning context $\mathbf{C}_k(t)$ is dynamically interpolated between the optimized variants and the original anchor based on the timestep. 
Finally, utilizing the \textit{Reference-Anchored Batching Strategy}, we compute rewards and update the policy.

\begin{algorithm}[h]
\caption{\ourshort Training Process}
\label{alg:ourshort}
\begin{algorithmic}[1]
\REQUIRE Pretrained policy $v^{\text{ref}}$, training policy $v_\theta$ initialized from $v^{\text{ref}}$;
\REQUIRE Reward function $R(\cdot)$, frozen text encoder $f_\phi(\cdot)$, prompt dataset $\mathcal{C}$;
\REQUIRE Semantic variants number $K$, diversity weight $\lambda_{\text{div}}$, perturbation range $\rho$.
\item[\textbf{Initialize:}] Data collection policy $v^{\text{old}} \leftarrow v^{\text{ref}}$, training policy $v_\theta \leftarrow v^{\text{ref}}$, data buffer $\mathcal{D} \leftarrow \emptyset$.
\FOR{each training iteration $i$}
    \STATE Sample batch of prompts $c \sim \mathcal{C}$
    \STATE \textcolor{gray}{// Embedding-Perturbed Mechanism}
    \FOR{each prompt $c$ in batch}
        \STATE Obtain static embeddings $\mathbf{E}$ and effective content index set $\mathcal{I}$
        \STATE Initialize perturbations $\Delta = \{\bm{\delta}_k\}_{k=1}^{K-1}$, where $\bm{\delta}_k \sim \mathcal{N}(0, \sigma_{\text{init}}^2 \mathbf{I})$
        \REPEAT
            \STATE Construct perturbed embeddings $\tilde{\mathbf{E}}_k$ via Eq.~\ref{eq:embedding_perturbed}
            \STATE Compute global embeddings $e_k = \operatorname{Pool}(f_\phi(\tilde{\mathbf{E}}_k))$ and anchor $e_{\text{anc}} = \operatorname{Pool}(f_\phi({\mathbf{E}}))$
            \STATE Calculate $\mathcal{L}_{\text{emb}} = \lambda_{\text{div}} \mathcal{L}_{\text{div}}(\Delta) + \mathcal{L}_{\text{anc}}(\Delta)$ using Eq.~\ref{eq:diversity} and Eq.~\ref{eq:similariyu}
            \STATE Update $\Delta \leftarrow \Delta - \alpha \nabla_\Delta \mathcal{L}_{\text{emb}}$
        \UNTIL{max steps}
        \STATE Freeze $\Delta$ and obtain optimized conditions $\{ \mathbf{C}^{\text{opt}}_{k} \}_{k=1}^{K-1}$
    \ENDFOR
    
    \STATE \textcolor{gray}{// Noise-Aware Sampling \& Reference-Anchored Batching}
    \FOR{each prompt $c$}
        \STATE Form batch $\mathcal{C}_{\text{batch}} = \{ \mathbf{C}^{\text{orig}} \} \cup \{ \mathbf{C}^{\text{opt}}_k \}_{k=1}^{K-1}$
        \STATE Sample $K$ images $\bm{x}_0^{1:K}$. During sampling, at timestep $t$:
        \STATE \quad Interpolate condition $\mathbf{C}_k(t) = \gamma(\sigma_t) \mathbf{C}^{\text{opt}}_k + (1 - \gamma(\sigma_t)) \mathbf{C}^{\text{orig}}$ (Eq.~\ref{eq:noise_aware})
        \STATE Evaluate rewards $r^{\text{raw}} = R(\bm{x}_0^{1:K},\mathbf{C}^{\text{orig}})$
        \STATE Define optimality probability: $r$ via Eq.~\ref{eq:r_normalization}
        \STATE Store tuple $\{c, \bm{x}_0^{1:K}, r^{1:K}\}$ in buffer $\mathcal{D}$
    \ENDFOR

    \STATE \textcolor{gray}{// Policy Optimization}
    \FOR{each mini-batch $\{c, \bm{x}_0, r\} \in \mathcal{D}$}
        \STATE Sample timestep $t$, noise $\bm{\epsilon}$. Forward process: $\bm{x}_t = \alpha_t \bm{x}_0 + \sigma_t \bm{\epsilon}$
        \STATE Implicit positive velocity: $\bm{v}_\theta^+(\bm{x}_t, \mathbf{C}^{\text{orig}}, t) := (1-\beta)v^{\text{old}}(\bm{x}_t, \mathbf{C}^{\text{orig}}, t) + \beta v_\theta(\bm{x}_t, \mathbf{C}^{\text{orig}}, t)$
        \STATE Implicit negative velocity: $\bm{v}_\theta^-(\bm{x}_t, \mathbf{C}^{\text{orig}}, t) := (1+\beta)v^{\text{old}}(\bm{x}_t, \mathbf{C}^{\text{orig}}, t) - \beta v_\theta(\bm{x}_t, \mathbf{C}^{\text{orig}}, t)$
        \STATE Update $\theta \leftarrow \theta - \lambda \nabla_\theta \left[ r \| \bm{v}_\theta^+ - \bm{v} \|_2^2 + (1-r) \| \bm{v}_\theta^- - \bm{v} \|_2^2 \right]$
    \ENDFOR
    \STATE Clear buffer $\mathcal{D} \leftarrow \emptyset$
\ENDFOR
\STATE \textbf{Output:} Optimized policy $v_\theta$
\end{algorithmic}
\end{algorithm}

\section{Experimental Setting Details}
\label{appendix:exp_details}
To ensure a fair comparison, we enforce a strict constraint on the same total optimization steps across all methods , ensuring that performance gains differ solely from algorithmic improvements rather than extended training duration.

\begin{figure}[!] %
    \centering
    \includegraphics[width=1.0\textwidth]{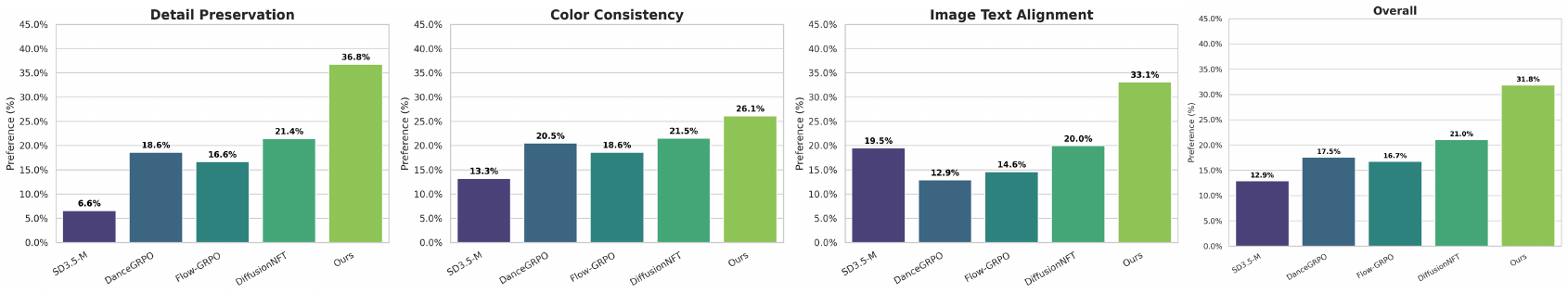} 
    \caption{\textbf{Human Preference Evaluation Results.} We compare \ourshort against SD3.5-M and RL-based baselines (DanceGRPO, Flow-GRPO, DiffusionNFT) across four key dimensions: Detail Preservation, Color Consistency, Image-Text Alignment, and Overall Quality. The results demonstrate that our method consistently achieves the highest user preference rates across all categories.}
    \label{fig:userstudy}
\end{figure}

\subsection{\ourshort Hyperparameters}
\label{appendix:our_hyperparams}

\ourshort introduces a decoupled exploration strategy defined by the latent group size $G$ (the number of different noises) and semantic variants $K$ (one unperturbed embedding and $K-1$ embedding perturbations). We tailor these hyperparameters to three distinct experimental configurations:
\begin{itemize}
    \item \textbf{High-Exploration Compositional Generation}: Optimized on GenEval with a larger semantic search space ($G=4, K=12$) for \textbf{300 steps}. This short schedule is sufficient to maximize reasoning capabilities given the dense signal provided by high-exploration variants, ensuring that the model's broad semantic coverage is preserved.
    \item \textbf{Efficient Compositional Generation}: Optimized on GenEval under a constrained compute budget ($G=2, K=4$) for \textbf{850 steps}. The slightly extended duration compensates for the smaller group size to demonstrate robust sample efficiency without compromising the learned diversity.
    \item \textbf{Human Preference Alignment}: Optimized on PickScore with a balanced configuration ($G=8, K=3$) for \textbf{750 steps}, prioritizing visual fidelity and precise alignment while maintaining generative diversity.
\end{itemize}

Detailed hyperparameters for the embedding-perturbed exploration, policy optimization and other specific algorithm settings are listed in Tab.~\ref{tab:full_hyperparams}.

\subsection{Details of Comparison with Baselines}
To ensure a rigorous comparison, we utilize \textbf{official open-source repositories} for all baselines, guaranteeing implementation correctness. 
While respecting original architectural configurations, we standardize reward settings and training steps across all methods. 
Moreover, consistent with the CFG-free design of \ourshort, we evaluate the SD3.5-M baseline without CFG to strictly isolate the improvements attributed to policy optimization.
Crucially, to ensure \textbf{optimization exploration parity}, we explicitly align the sampling batch size of baseline methods with the number of semantic variants $K$ employed in \ourshort, ensuring that all methods operate under an identical search budget.

\textbf{Evaluation Protocols.}
Building on this standardized setup, we design specific evaluation protocols tailored to the distinct goals of each experimental setting:

\begin{itemize}
    \item \textbf{Protocol for Compositional Reasoning (GenEval):} In this setting, the primary objective is to verify precise instruction adherence while maintaining generative breadth. Therefore, we assess performance using the \textbf{GenEval} score as the direct task metric, alongside \textbf{DivGenBench} metrics (IDS, ASC, SDI, PVS) to monitor diversity.
    
    \item \textbf{Protocol for Human Preference Alignment (PickScore):} In preference optimization, a key risk is "gaming" the specific reward model without genuine quality improvement. Consequently, beyond the direct \textbf{PickScore} objective and \textbf{DivGenBench} diversity metrics, we explicitly evaluate \textbf{Transferability} on unseen reward models (Aesthetic Score, ImageReward, and HPSv2.1). This ensures that the optimized policy achieves robust visual quality improvements rather than overfitting to the proxy reward.
\end{itemize}

\begin{table}[t]
\centering
\caption{\textbf{Hyperparameter Specifications.} Default settings used for training \ourshort across all experiments unless otherwise stated.}
\label{tab:full_hyperparams}
\resizebox{0.8\linewidth}{!}{
    \setlength{\tabcolsep}{5pt} 
    \renewcommand{\arraystretch}{1.15} 
    \begin{tabular}{lc | lc | lc}
    \toprule
    \multicolumn{2}{c|}{\textbf{General Settings}} & \multicolumn{2}{c|}{\textbf{Embedding-perturbed Exploration}} & \multicolumn{2}{c}{\textbf{Policy Optimization}} \\
    \midrule
    \textbf{Parameter} & \textbf{Value} & \textbf{Parameter} & \textbf{Value} & \textbf{Parameter} & \textbf{Value} \\
    \midrule
    Random Seed & 42 & Optimize Steps ($T_{\text{emb}}$) & 300 & Policy LR ($\eta$) & \num{3e-4} \\
    Optimizer & AdamW & Embedding LR ($\alpha_{\text{emb}}$) & \num{1e-3} & Weight Decay ($\lambda_{\text{wd}}$) & \num{1e-4} \\
    Mixed Precision & fp16 & Init. Std Dev ($\sigma_{\text{init}}$) & \num{1e-4} & KL Penalty ($\beta_{\text{kl}}$) & \num{1e-4} \\
    Distributed Mode & DDP & Max Norm Clip ($\|\delta\|_{\max}$) & 0.05 & Grad Clip Norm & 1.0 \\
    LoRA Rank ($r$) & 32 & Diversity Weight ($\lambda_{\text{div}}$) & 50.0 & Grad Accumulation Per Epoch & 1 \\
    LoRA Alpha ($\alpha$) & 64 & Anchor Target ($\mu$) & 0.80 & Train Sample Steps & 10 \\
    Resolution & $512^2$ & Anchor Margin ($\epsilon$) & 0.01 & Train Guidance & 1.0 \\
    Use EMA & True & Perturbation Range ($\rho$) & 0.4 & Inference Steps & 40 \\
    \bottomrule

    \end{tabular}
}
\end{table}

\section{Extended Experiments}

\subsection{User Study}
\label{appendix:user_study}

To validate the effectiveness of \ourshort in alignment with human preferences, we conducted a comprehensive user study following the protocol established in \cite{dancegrpo}. We compared our method against the backbone model Stable Diffusion 3.5 Medium (SD3.5-M)  and three competitive state-of-the-art RL-based baselines: Flow-GRPO \cite{flowgrpo}, DanceGRPO\cite{dancegrpo} and DiffusionNFT \cite{diffusionnft}.

To ensure a rigorous evaluation, we randomly sampled prompts from the PickScore validation set. For each prompt, anonymized images generated by the five methods (SD3.5-M, DanceGRPO, Flow-GRPO, DiffusionNFT, and \ourshort) were presented to human evaluators in a side-by-side comparison format. The evaluators were asked to select the best image based on four distinct criteria established in \cite{dancegrpo}.

The results of the user study are illustrated in Fig\ref{fig:userstudy}. \ourshort consistently outperforms the baselines across all evaluated dimensions. Notably, our method achieves a significant lead in \textit{Image-Text Alignment} with a preference rate of 33.1\% and dominates in \textit{Overall Quality} with 31.8\%, compared to the competitive baselines. This human preference data corroborates our quantitative findings, confirming that the embedding-perturbed exploration strategy effectively mitigates reward hacking, resulting in generation trajectories that are more aligned with human intent.

\subsection{Additional Ablation Study}
\label{appendix:ablation}
\paragraph{Ablation of Perturbation Scope.} 
\begin{figure}[!] %
    \centering
    \includegraphics[width=0.65\linewidth]{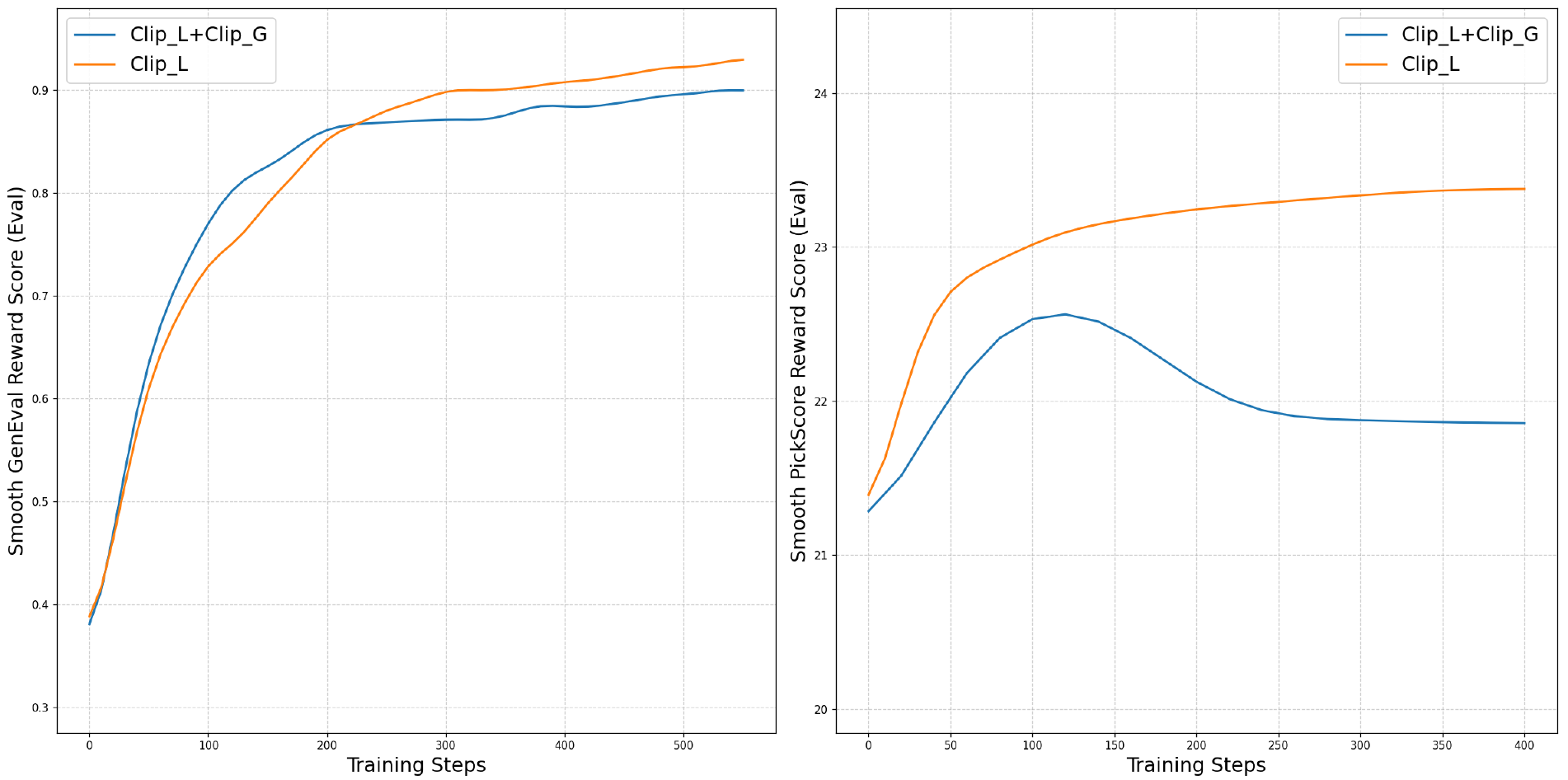} 
    \caption{\textbf{Ablation of Perturbation Scope.} We compare perturbing the primary encoder (CLIP-L) against perturbing both encoders. The plot shows that limiting perturbation to CLIP-L ensures high-quality generation, while the dual-encoder strategy suffers from significant performance degradation and instability.}
    \label{fig:ablation_clip}
\end{figure}
We compare the impact of perturbing only the primary encoder (CLIP-L) versus both encoders (CLIP-L + OpenCLIP-G) in Fig.~\ref{fig:ablation_clip}. 
The results show that targeting CLIP-L alone consistently yields superior and more stable performance. 
In contrast, perturbing both encoders simultaneously leads to severe instability and performance collapse, particularly in PickScore optimization. 
This suggests that expanding the perturbation space to both encoders increases optimization complexity and makes it difficult to bound semantic drift. 
Restricting intervention to CLIP-L provides the best balance, offering enough flexibility for diversity while ensuring stable and high-quality generation.

\paragraph{Ablation of Coefficient $\lambda_{div}$.}
\begin{table}[t]
    \centering
    \footnotesize 
    \caption{\textbf{Ablation of Coefficient $\lambda_{div}$.} We evaluate the impact of $\lambda_{div}$ to determine the optimal configuration. We find that $\lambda_{div}=50$ yields peak performance for PickScore and dense exploration ($K=12$) on GenEval, while sparse candidate sets ($K=4$) on GenEval benefit from a larger $\lambda_{div}$.}
    \label{tab:ablation_div}
    \begin{tabular*}{0.55\linewidth}{@{\extracolsep{\fill}}lcccc}
        \toprule
        \multirow{2}{*}{$\lambda$} & \multicolumn{2}{c}{GenEval} & & PickScore \\
        \cmidrule(lr){2-3} \cmidrule(lr){5-5}
         & $K=12$ & $K=4$ & & $K=3$ \\
        \midrule
        30  & 0.924 & 0.902 & & 23.14 \\
        50  & \textbf{0.932} & 0.916 & & \textbf{23.38} \\
        100 & 0.930 & \textbf{0.917} & & 22.86 \\
        \bottomrule
    \end{tabular*}
    
\end{table}
In Tab.~\ref{tab:ablation_div}, we evaluate how the repulsive coefficient $\lambda_{div}$ balances exploration with semantic fidelity.
The results show that the optimal $\lambda_{div}$ is governed by the exploration density and task characteristics. 
For GenEval, sparse candidate sets ($K=4$) require a larger $\lambda_{div}$ to escape local modes, whereas dense sets ($K=12$) favor a smaller penalty to avoid semantic degradation caused by excessive repulsion. 
In contrast, PickScore optimization, which focuses on fine-grained aesthetic alignment rather than drastic structural shifts, achieves a stable equilibrium at a moderate $\lambda_{div}=50$, balancing visual diversity with generative fidelity.

\section{Visualization}
\label{appendix:visualization}
\begin{figure}[!] %
    \centering
    \includegraphics[width=0.85\linewidth]{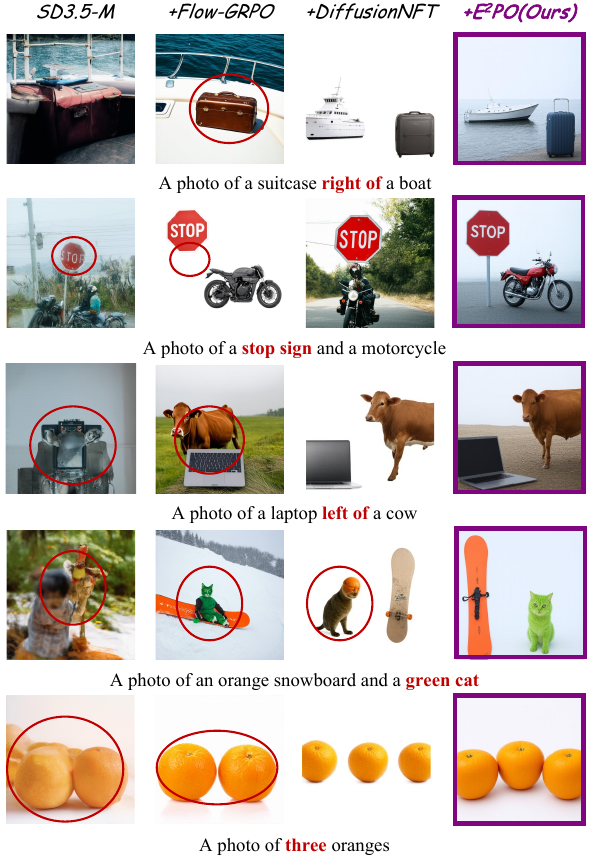} 
    \caption{\textbf{Qualitative Comparison of \ourshort against SOTAs on the GenEval Benchmark.} All RL-based methods are trained using GenEval as the reward model.}
    \label{fig:visualization_appendix_geneval}
\end{figure}

\begin{figure}[!] %
    \centering
    \includegraphics[width=0.85\linewidth]{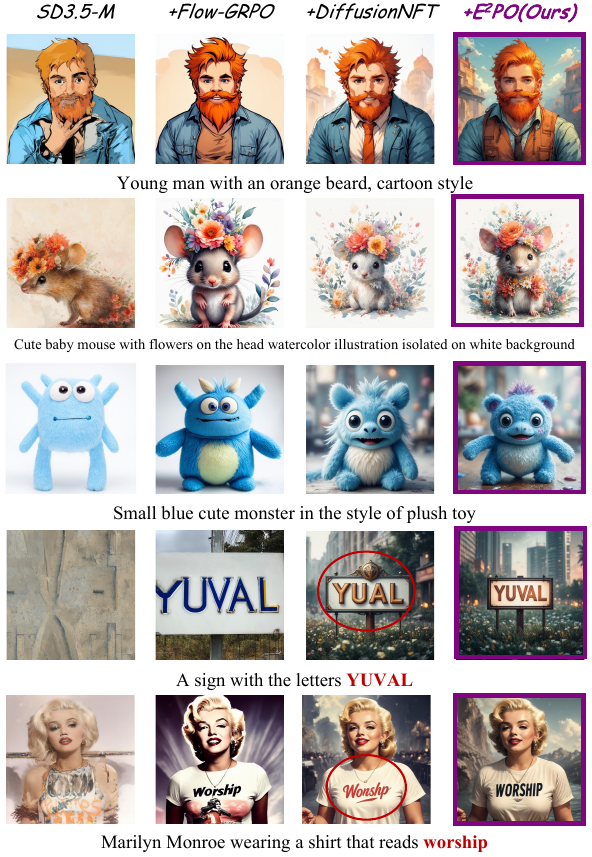} 
    \caption{\textbf{Qualitative Comparison of \ourshort against SOTAs on the PickScore Benchmark.} All RL-based methods are trained using PickScore as the reward model.}
    \label{fig:visualization_appendix_pickscore}
\end{figure}

In this section, we provide additional visualization results to assess the generation quality of \ourshort. We conduct qualitative comparisons against SOTA baselines on two distinct benchmarks: GenEval and PickScore. Fig.~\ref{fig:visualization_appendix_geneval} displays the results for GenEval, while Fig.~\ref{fig:visualization_appendix_pickscore} presents the comparison on PickScore. In both scenarios, to ensure a consistent evaluation, all RL-based methods were trained using the respective benchmark metric (GenEval or PickScore) as the reward signal.

\end{document}